\documentclass[11pt,a4paper]{article}
\usepackage[hyperref]{acl2019}
\usepackage{times}
\usepackage{latexsym}
\usepackage{verbatim}
\usepackage{xspace}
\usepackage{booktabs}
\usepackage{graphicx}
\usepackage{amssymb}
\usepackage{mathtools}

\usepackage{adjustbox}
\usepackage{array}

\newcolumntype{R}[2]{%
    >{\adjustbox{angle=#1,lap=\width-(#2)}\bgroup}%
    l%
    <{\egroup}%
}

\usepackage{url}

\aclfinalcopy

\mathtoolsset{showonlyrefs}

\title{Zero-Shot Fine-Grained Style Transfer:
Leveraging Distributed Continuous Style Representations to Transfer 
To Unseen Styles}

\author{
  \qquad \quad Eric Michael Smith, Diana Gonzalez-Rico, Emily Dinan, Y-Lan Boureau\\
  \qquad \quad Facebook AI Research \\
}

\date{}

\begin{document}
\maketitle

\begin{abstract}
Text style transfer is usually performed using attributes that can take a handful of discrete values (e.g., positive to negative reviews). In this work,
we introduce an architecture that can leverage pre-trained consistent continuous
distributed style representations and use them to transfer
to an attribute unseen during training, without requiring any
re-tuning of the style transfer model.
We demonstrate the method by training an architecture to transfer text conveying one
sentiment to another sentiment, using a fine-grained set of over 20 sentiment labels rather than the binary positive/negative often used in style transfer. Our experiments show that
this model can then rewrite text to match a target sentiment that was
unseen during training.
\end{abstract}

\section{Introduction}
A time-honored way to nudge human creativity is to structure generation around
the idea of variation, from literary pastiches to variations in classical
music or the concept of jazz standards. Variation is
then used primarily as an inspiration device, where it is not necessary to stick too closely to the original template.
Artificial text style transfer can similarly act as a loosely constrained generative device,
to combat monotony by generating more
variations of a given piece of text, or to avoid blandness through
anchoring on an interesting original.
Within that framing, it is more important to be able to generate 
richer variations than to strictly preserve content.

Most existing text style transfer work has focused on a narrow set of
applications where the attributes of interest have a very limited
set of discrete possible values, e.g. two valences of reviews (positive and negative), three different writing styles [example], five types of
restaurant cuisines \citep{lample2018multipleattribute}.
This is very well suited to applications where style transfer
has to adhere closely to its input (e.g., editing text to make it more formal or business-like), but less so when the emphasis is on creativity
more than faithfulness to the original.
In this work, we propose a new approach that allows for text generation
conditioned on a much richer and fine-grained specification of target
attributes, by leveraging distributed representations pre-trained through a separate supervised classification task.
By specifying attributes through continuous distributed representations, we show that our architecture allows for fine-grained conditioned text generation that can match new attribute targets unseen during training, or attribute targets implicitly specified through text, that may not precisely match any of the discrete labels originally used to define the attribute space.

This work thus makes the following contributions: first, we propose a
method that allows transfer to a much larger set of fine-grained
styles without requiring additional optimization during inference.
Second, we show how this method can be used to perform zero-shot style
transfer to new styles unseen during the style transfer training,
through leveraging a joint underlying lower-dimensional style embedding space.
Third, we show how fine-tuning a pre-trained attribute control architecture affords control over a different but related attribute space.

\section{Related work}
Many earlier approaches to text style transfer rely on a disentangling objective seeking to extract a representation from which the original style is hard to recover \citep{lample2017fader}.
However, recent work has shown that this disentanglement was neither empirically achieved, nor necessary \citep{lample2018multipleattribute}. In this work, we do not use any disentanglement objective either.

Style transfer can be viewed as translation from one style to another. Recent strides in unsupervised translation have led to a body of work adapting machine translation techniques to style transfer \citep{prabhumoye2018style,lample2018multipleattribute,zhangstyletransfer}. This work follows this approach and uses an architecture very similar to that in \citet{lample2018multipleattribute}.

When used to generate a richer set of alternatives, style transfer can be viewed as a controlled text generation technique with a particularly 
strong conditioning anchor. The recently released CTRL model \citep{keskar2019ctrl}
allows for generation based on control codes such as a specific website link, which are used as a pre-pended token. The style attribute is similarly specified here by providing an initial token to the model to specify the target attribute, but the generated text is also conditioned much more strongly on a source sentence, as was done in \citet{lample2018multipleattribute}.

There has been recent work on achieving fine-grained graded style transfer by editing the hidden representation of an input towards one that would be classified more readily into a target style \citep{wang2019controllable,liu2019revision}, or sampling responses around a given output to select those that better match a target style \citep{gao2019structuring}.
These methods can be viewed as a positive version of the disentangling methods that were leveraging an adversarial classifier to prevent classification into the source attribute, instead pushing the hidden representation towards classification into the target attribute.

In this work, we instead propose to decouple the classifier from the style transfer architecture by merely using the classifier to produce a distributed representation of the target attribute, so that existing pre-trained supervised representations can be re-used. This would allow for our method to be applied to any type of consistent distributed embedding
space (e.g., pre-trained unsupervised fastText embeddings~\citep{joulin2016bag}).

\section{Specifying target attributes as distributed continuous representations}
Our approach relies on an autoencoder architecture similar to that in \citet{lample2018multipleattribute}, modified to leverage
consistent pre-trained distributed continuous representations of attributes. This section presents the notation and base architecture before introducing our key modification to leverage embeddings.

\subsection{Base architecture}
 
This section briefly introduces the architecture and training objective of \citet{lample2018multipleattribute}, which we use as base for our style transfer system.

 Let $\mathcal{D}={(x^i,y^i)}_{i \in [1,n]}$ be a training set of $n$ sentences $x^i \in \mathcal{X}$ paired with source attribute values $y^i$.
$y^i \in \mathcal{Y}$ is a discrete attribute value in the set $\mathcal{Y}$ of possible values for the attribute being considered, e.g. $\mathcal{Y}=\{\text{bad},\text{neutral},\text{good}\}$ if $y^i$ represents the overall rating of a restaurant review. In this work, we only consider transfer of a single attribute, but our approach could easily be extended to multiple attributes using an attribute embedding averaging heuristic as in \citet{lample2018multipleattribute}.

The style transfer architecture consists of a model $F : \mathcal{X} \times \mathcal{Y} \rightarrow \mathcal{X}$ that maps any pair $(x,\tilde{y})$ of a source sentence $x$ (whose source attribute is $y$) paired with a target attribute $\tilde{y}$ to a new sentence $\tilde{x}$ that has the target attribute value $\tilde{y}$, while striving to remain as close as possible to $x$, and being fluent English. This is achieved by training a sequence-to-sequence auto-encoder as a denoising auto-encoder, with an added back-translation objective to ensure transfer to the target attribute. 

The input $x$ is encoded into a latent representation $z = e(x)$,  then $(z, \tilde{y})$ is decoded into $\tilde{x} = d(z, \tilde{y})$, where the parameters of encoder $e$ and decoder $d$ are trainable, and target attribute value $\tilde{y}$ can be a different attribute -- or the same original attribute if not trying to modify it when reconstructing.

\paragraph{Denoising objective}

In order to retain fluency and ability to reconstruct well without
merely copying, the architecture is trained with a denoising auto-encoding objective $L_{AE}$~\citep{fu2017style}:

\begin{equation}
    L_{AE} = \sum_{(x,y) \sim \mathcal{D}} - \log p_d \Big(x | e(x_c), y\Big),
\end{equation}

where $x_c$ is a noisy version of input text $x$ corrupted with
word drops and word order shuffling as described in~\citet{lample2017unsupervised} and $p_d$ is the probability distribution over sequences $x$ induced by the decoder. Here, the input is reconstructed without changing the source attribute value.

\paragraph{Back-translation objective}

The decoder is encouraged to leverage the provided target attribute
through a back-translation loss~\citep{sennrich2015improving, lample2017unsupervised, lample2018phrase, unsupNMTartetxe}:
input $x$ is encoded into $z$, but then decoded using 
target attribute value $\tilde{y}$, yielding the reconstruction $\tilde{x}$. $\tilde{x}$ is in turn used as input of the encoder and decoded using the source attribute value $y$ to ideally obtain the source $x$, and we train the model to map $(\tilde{x}, y)$ back into $x$.
The back-translation objective $L_{BT}$ is thus written:

\begin{equation}
    L_{BT} = \sum_{\mathclap{(x,y) \sim \mathcal{D}, \tilde{y} \sim \mathcal{Y}}}  - \log p_d \bigg(x | e\Big( d\big(e(x), \tilde{y}\big) \Big), y\bigg),
\end{equation}

where $d(e(x), \tilde{y})$ is a variation of the input sentence $x$ written with a randomly sampled target attribute $\tilde{y}$ that
is specified according to the procedure described in sec.~\ref{sec:cont}.
Back-translated sentences are generated on the fly during training
by greedy decoding at each time step.

\paragraph{Overall objective}

The system is trained by combining both denoising auto-encoding
and back-translation loss:
\begin{equation}
    \mathcal{L} = \lambda L_{AE} + (1 - \lambda) L_{BT},
\end{equation}

where the mixture hyperparameter $\lambda$ is optimized over the validation set to achieve the best combinations of the metrics specified below, as in \citet{lample2018multipleattribute}.
We optimize this loss by stochastic gradient descent without back-propagating through the back-translation generation process.

\paragraph{Architecture building blocks} The encoder $e$ is a 2-layer bidirectional LSTM using word embedding look-up tables trained from scratch. The decoder $d$ is a 2-layer LSTM augmented with an attention mechanism~\citep{bahdanau2014neural}. All the embedding and hidden layer dimensions are 512, including the attribute embedding obtained as explained in Section~\ref{sec:cont}. Decoding is conditioned on both that attribute embedding, which is provided as the first token embedding, similar to \citet{lample2018phrase}, and on a representation of the input obtained from the encoder with an attention mechanism. 

\subsection{Leveraging pre-trained distributed continuous representations}
\label{sec:cont}
\citet{lample2018multipleattribute} specify the target attribute
as an embedding read from a lookup table that is optimized during training. This means that each target attribute value has its own entry,
and precludes leveraging known similarities between target attribute values.

Instead, we propose to write the target embedding $y = Wy_d$ as the product
of an existing distributed embedding $y_d$, and a weight matrix
$W$. The motivation for this is that pre-trained distributed embeddings encode similarities between attribute values that can be learned from other tasks (e.g., supervised classification) and directly leveraged for style transfer.

In this work, we obtain the embedding by running some text $\hat{x}$ possessing the desired target attribute value through a feedforward classifier $y_d = c(\hat{x})$.
We experiment with a fastText classifier \citep{joulin2016bag} and
a classifier derived from BERT \citep{devlin2018bert} with an added bottleneck layer, and use the last hidden layer whose dot-product with class embeddings would 
determine what class is selected. The dimension of that layer is arbitrary. Preliminary experiments have shown better training with smaller dimensions, so in the remainder of the paper we set the supervised embedding dimension to 8. Thus, the weight matrix $W$ is of dimension $512 \times 8$. Note that the base style transfer architecture adapted from \citet{lample2018multipleattribute} for $k$ possible attribute values would correspond to $W$ being a look-up table of dimension $512 \times k$, with a one-hot encoding of each attribute value instead of the supervised distributed embeddings used here.

During training, randomly selected samples from the training set are run through the classifier to obtain a fine-grained continuous distributed target embedding value which is used as target attribute value for the back-translation loss, and scaled to unit norm. For validation and measuring accuracy of transfer, class embeddings are used instead, after being also scaled to unit norm.

\section{Experiments in original fine-grained attribute space}
We demonstrate the technique using a set of fine-grained sentiment labels such as happy, curious, angry, hopeful, sad, thankful, etc. (see full list in Table~\ref{tab:label_list}). 
\begin{table}
\centering
\begin{small}
\begin{tabular}{p{1.5cm}p{5.5cm}}
\toprule
Base task & aggravated, angry, annoyed, confused, curious, 
\textit{delighted}, ecstatic, \textit{emotional}, fabulous, fantastic, 
frustrated, grateful, happy, heartbroken, hopeful, 
\textit{irritated}, joyful, overwhelmed, \textit{perplexed}, pumped, 
sad, shocked, sleepy, thankful \\
\midrule
ED task & afraid, angry, annoyed, anticipating, anxious, apprehensive, ashamed, caring, confident, content, devastated, disappointed,  disgusted, embarrassed, excited, faithful, furious, grateful, guilty, hopeful, impressed, jealous, joyful, lonely, nostalgic, prepared, proud, sad, sentimental, surprised, terrified, trusting \\
\bottomrule
\end{tabular}
\end{small}
\caption{Top: set of 24 sentiment labels used as attribute values for training of the style transfer architecture. Experiments in Section~\ref{sec:fine-grained} train architectures to transfer between all 24 labels and show good transfer performance (see
Table~\ref{tab:none_held_out_auto_metrics}). Experiments in Section~\ref{sec:zero_shot} use 20 for training the style transfer architecture, while the four labels shown in italics are not seen during training, but still obtain reasonable transfer performance,
as seen in Table~\ref{tab:some_held_out_auto_metrics}.
Bottom: set of 32 labels used in the \textsc{EmpatheticDialogues}
dataset. Experiments exploring transfer to that space are described in Section~\ref{sec:ed} with results shown in Table~\ref{tab:ed_auto_metrics}.
}
\label{tab:label_list}
\end{table}
The choice of fine-grained sentiment as set of attributes is motivated by the
richness of the attribute space, for which large labelled datasets 
are available (e.g., \citet{DailyDialog,rashkin2019empathy}), while also being in continuity with the use of sentiment as style in much of the text style transfer literature.

\subsection{Dataset}
We train a sentiment classifier over 24 sentiments using an unreleased dataset
of millions of samples of social media content written by English speakers
with a writer-assigned sentiment tag. In order to make our work reproducible by others, we select training data from publicly available data in the following way: starting from a Reddit dump collected and published by a third party, we use that classifier
to select a subset of millions of posts matching
each of the 24 sentiment labels of interest.
A new classifier is
then trained from scratch on that data to provide the target embeddings, and the initial classifier is discarded.
We pick a set of 24 sentiment labels to demonstrate fine-grained transfer to a larger set of possible labels compared to previous work, which usually limits transfer to a handful of possible attribute values. The set of 24 sentiment labels (see Table~\ref{tab:label_list}) is selected by keeping sentiment labels that have reasonable-looking matches among the Reddit posts from the third-party dump, after a quick manual inspection of random samples to determine which labels to keep and what threshold to use to decide which posts to retain. Posts from the third-party Reddit dump that score above those thresholds are run through the safety classifier
from \citet{dinan2019safety} to remove offensive or toxic content, and the English language classifier from fastText \cite{joulin2016bag} to remove non-English content. We also
remove content that contains URLs or images.
The remaining data comprises between 22k and 11M examples per sentiment label, and data from each label is sampled in a balanced way during training. The final data consists of a train set of 31M labeled samples, and an additional 730k samples as validation and test sets, respectively.

\subsection{Evaluation}

Following \citet{lample2018multipleattribute}, we use three
automated metrics to measure target attribute control, fluency, and content preservation:
\begin{itemize}
    \item \textbf{Attribute control: } Attribute control is measured by using a fastText or BERT classifier trained to predict attribute values. This classifier does not have the low-dimensional bottleneck of the one used to produce the  embedding $y_d$, as classification performance is more accurate with larger dimensions. 
    \item \textbf{Fluency: } Fluency is measured by the perplexity assigned to generated text sequences by an LSTM language model trained on the third-party Reddit training data.
    \item \textbf{Content preservation: } Content preservation is roughly captured through n-gram statistics, by measuring the \textit{BLEU} score between generated text and the input itself (called \textit{self-BLEU} as in \citet{lample2018multipleattribute}).
\end{itemize}

The best trade-off between those three aspects of transfer is dependent on the desired application. If the goal is to generate new utterances for a retrieval system in a conversation while keeping them from being bland or too repetitive through anchoring on a source utterance, in a manner reminiscent of the retrieve-and-refine approach \citep{weston2018retrieve}, fluency and attribute control would matter more than content preservation. If the goal is to stick as close to the source sentence as possible and say the same things another way, which is better defined for language types (e.g., casual vs. formal) than for sentiment, then content preservation would matter more, but in a way that self-BLEU might not be sophisticated enough to capture.

Hyperparameters are picked by looking at performance over the validation set, using self-BLEU and transfer control. We also experimented with pooling (as in \citet{lample2018multipleattribute}) and sampling with a temperature instead of greedy decoding, as well as larger bottleneck dimensions, but these all resulted in worse performance on the datasets we use here.
Evaluation is performed by running style transfer on all non-matching combinations of source and target labels, on up to 900 source sequences per source label.
Results are reported using source sentences from the test set.

\begin{table}
\begin{small}
\begin{tabular}{lp{6cm}}
\toprule
\textbf{source} & \textbf{it is annoying how Meme has already changed meanings...
} \\
Model 2 & it is fantastic football Meme has already changed meanings...
 \\
Model 4 & it is fantastic =D
 \\
\midrule
\textbf{source}& \textbf{I wish people would stop making right-handed Link pics.
} \\
Model 2 & Fantastic show in right-handed Link pics.
 \\
Model 4 & I think this is fantastic and Star Wars videos...
 \\
\bottomrule
\end{tabular}
\end{small}
\caption{Generations from models 2 and 4 in Table~\ref{tab:none_held_out_auto_metrics}, transferring from {\em annoyed} to {\em fantastic}. Different stages in the training lead to different trade-offs between attribute control, content preservation, and fluency: model 2 preserves a lot more of the source sentence, while model 4 has better attribute control but retains little from the source sentence.}
\label{tab:model_gen}
\end{table}

\subsection{Fine-grained style transfer}
\label{sec:fine-grained}
We first use our system to demonstrate successful transfer over
a large number of fine-grained attribute values.
Results in Table~\ref{tab:none_held_out_auto_metrics} show
that training achieves very good accuracy while maintaining
reasonable self-BLEU scores and perplexity similar to the average
perplexity of reference sentences. Classification of the identity baseline to the source attribute is a bit less than classification to the target attribute for the target baseline because the former uses test set examples, which were not seen by the classifier.
Example generations are given in Table~\ref{tab:some_held_out_generations}, where four sentiment classes are held-out during training, but training is otherwise similar.

\begin{table}
    \centering
    \begin{small}
    \begin{tabular}{p{2.5cm}rrrr}
\toprule
& \multicolumn{2}{c}{Classification} &  &  \\
\cmidrule(lr){2-3}
& Target & Source & self-BLEU  & PPL \\
\midrule
Identity & 0.3 & 93.7 & 100.0 & 146.8 \\
Target attr. sample & 99.8 & 0.0 & 0.0 & 151.2 \\
\midrule
Model 1& 84.2 & 7.2 & 42.8 & 261.1 \\
Model 2& 91.0 & 3.6 & 36.8 & 225.7 \\
Model 3& 93.1 & 2.5 & 31.5 & 212.8 \\
Model 4& 97.1 & 0.5 & 6.0 & 129.7 \\
\bottomrule    
\end{tabular}
    \end{small}
    \caption{\label{tab:none_held_out_auto_metrics}
    Automated metrics on the fine-grained sentiment
    transfer task over 24 possible labels. Results are averaged over all transfer directions. Classification metrics show
    percentage of the generations classified as Target and Source
    label attributes. Successful sentiment transfer shifts classification from Source to Target attribute. Self-BLEU measures closeness to the source sequence.
    Perplexity (PPL) probes fluency.
    Top two rows show two trivial baselines: {\em Identity} copies the source sequence and gives the baseline no-transfer test-set metrics, and has minimal classification as the Target class. 
    {\em Target attr. sample} uses a random example from the target category training set as generation.
    Models 1 to 4 show different stages of the training, showing that different trade-offs between the three objectives of
    content preservation, attribute control and fluency can be achieved. Example generations for models 2 and 4 are shown in Table~\ref{tab:model_gen}.
    }
\end{table}

\begin{table}
\begin{small}
\begin{tabular}{lp{5.4cm}}
\toprule
\textbf{grateful} & \textbf{I appreciate him. And I love him.} \\
angry & I hate him. And I am angry about him. \\
hopeful & I would love him. And I hope it's true. \\
sad & I miss him. And I liked him. \\
thankful & I have seen him. And thanks for doing that. \\
\midrule
\textbf{hopeful} & \textbf{I hope I'm not too late to the party.} \\
angry & I am so angry I'm not too late to the party. \\
curious & I wonder if I'm not too late to the party. \\
ecstatic & I am ecstatic I'm not too late to the party. \\
happy & I am happy I'm not too late to the party. \\
\midrule
\textbf{pumped} & \textbf{Thank you! So pumped to pick this up!} \\
curious & Am I the only one who didn't pick this up? \\
frustrated & Of course it would be hard to pick this up! \\
hopeful & Any chance I can pick this up? \\
\midrule
\textbf{shocked}& \textbf{But she was shocked when she found out what'd happened.} \\
angry & But she was so angry when she found out what'd happened. \\
curious & Do you know if she found out what'd happened. \\
\textit{delighted} & \textit{Hey she laughed when she found out what'd happened.} \\
ecstatic & Absolutely ecstatic when she found out what'd happened. \\
\textit{emotional} & \textit{But she cried when she found out what'd happened.} \\
thankful & Thank you, she was looking forward to something like what'd happened. \\
\bottomrule
\end{tabular}
\end{small}
\caption{Example transfer generations from sequences from the test set of the third-party Reddit data, with various source sentiment labels (bold),
to various fine-grained target sentiment labels.
The bottom cell includes transfer to held-out labels that were not seen during training, in italics. Generations are from the model shown in the top row of Table~\ref{tab:some_held_out_auto_metrics}.}
\label{tab:some_held_out_generations}
\end{table}

\subsection{Zero-shot style transfer to unseen attribute values}
\label{sec:zero_shot}
Limiting the capacity of the attribute value representations through a small-dimensional bottleneck may make it easier for the auto-encoder to learn to generalize over the embedding space overall, beyond the specific combinations of the sentiment labels seen during training. 
To check if the transfer can indeed generalize to unseen sentiment labels, we train a system with 20 out of the 24 sentiment labels, holding out 4 labels that are seen by the classifier (shown in italics in Table~\ref{tab:label_list}),
but not the style-transfer auto-encoder architecture during training.
We then evaluate transfer to these unseen classes.
Results in Table~\ref{tab:some_held_out_auto_metrics} show that transfer to these unseen classes
is still largely successful, with the target class being picked more
than half the time out of 24 possible classes. However,
transfer to these held-out classes remains less successful than transfer to the classes seen during training. Examples of transfer to unseen classes are given at the bottom of Table~\ref{tab:some_held_out_generations}.

\begin{table}
    \centering
    \begin{small}
    \begin{tabular}{rrrrrrrr}
\toprule
\multicolumn{4}{c}{Training target attribute}
&
\multicolumn{4}{c}{Held-out target attribute} \\
\cmidrule(lr){1-4}
\cmidrule(lr){5-8}
\multicolumn{2}{c}{Classification} &  &  &
 \multicolumn{2}{c}{Classification} &  &  \\
\cmidrule(lr){1-2} 
\cmidrule(lr){5-6}
 Target & Sce & s-BL & PPL
&  Target & Sce & s-BL & PPL \\
\midrule
86.8 & 6.0 & 39.5 & 257.2
& 56.5 & 11.6 & 40.2 & 283.9 \\
90.5 & 4.2 & 36.7 & 240.5
& 62.2 & 9.2 & 38.5 & 285.5 \\
92.6 & 2.8 & 29.7 & 212.4
& 63.4 & 7.5 & 32.3 & 272.6 \\
\bottomrule    
\end{tabular}
    \end{small}
    \caption{Evaluation when 4 out of the 24 sentiment labels are held out during training, shown for three different stages of the training which capture three different trade-offs between the criteria of attribute control, content preservation, and fluency. The metrics shown are the same as in Table~\ref{tab:none_held_out_auto_metrics}: percentage classifications assigned to the target and source (Sce) attributes, self-BLEU (s-BL), and perplexity (PPL). 
    Left: transfer to target attributes seen by the style transfer architecture during training. Metrics are very similar to those obtained when training on 24 classes, in Table~\ref{tab:none_held_out_auto_metrics}. 
    Right: transfer to the 4 unseen classes is still largely successful, with
    the target attribute being selected more than half the time out of 24 possible attributes (chance would be 4\%), but clearly less so than for the attributes seen during training. S-BL scores are similar to those of attributes seen during training, but PPL is higher.}
    \label{tab:some_held_out_auto_metrics}
\end{table}

\section{Transferring to a new, related attribute space}
\label{sec:ed}

Training the style transfer architecture requires millions of training examples. In this section, we examine whether it is possible to leverage pre-training on a given sentiment transfer task, to then transfer\footnote{Note that transfer in this sentence is used first in the context of transfer learning, then in the context of style transfer.} that training to an attribute transfer task with a training set orders of magnitude smaller, as long as the attribute space is related.

\subsection{Dataset}
The dataset we use here to examine transfer to a related task is the \textsc{EmpatheticDialogues} dataset \citep{rashkin2019empathy}, which comprises about 25k dialogues accompanied by a situation description of a few sentences, and a sentiment label belonging to a list of 32, some of which are also in the list of 24 from the first task (e.g., angry, grateful, joyful, as shown in Table~\ref{tab:label_list}). We use the situation descriptions and sentiment labels, not the dialogues.

We perform evaluation using the same metrics as before.
The classification task over the \textsc{EmpatheticDialogues} labels is overall more difficult, given that there are more labels, but more importantly, that the dataset has not been pre-filtered by a classifier in the same way that the base training dataset was selected from the third-party Reddit dump. Thus, classification metrics (shown in Table~\ref{tab:ed_auto_metrics}) are lower across the board, with the upper bound being the 56.5\% of the Source classification for the Identity baseline. The language in \textsc{EmpatheticDialogues} is also easier to predict than that of Reddit, resulting in lower perplexity scores.

\begin{table}
\begin{small}
\begin{tabular}{lp{5.8cm}}
\toprule
\textbf{source} & \textbf{I come home from work and my parents are always arguing. It frustrates me.
} \\
Scratch & I have a big presentation at work that I am really looking forward to it.
 \\
Zero-shot & I come home from her and my parents are always arguing. It compliments me.
 \\
 Fine-tuned & I come home from work and my parents are always studing. I am so content with my wife. \\
\midrule
\textbf{source}& \textbf{My boss made me work overtime yesterday and I didn't even get paid for it!
} \\
Scratch & My husband and I went on a vacation trip to New York. I was not expecting it
 \\
Zero-shot & My boss made it overtime kicked and I didn't even get arrested for it!
 \\
 Fine-tuned & My boss made me work yesterday. Everything I had is going well now. \\
\bottomrule
\end{tabular}
\end{small}
\caption{Generations from various transfer methods to perform attribute control over \textsc{EmpatheticDialogues}, with models from Table~\ref{tab:ed_auto_metrics}, rewriting from {\em annoyed} to {\em content}. Training from scratch mostly ignores source content. Zero-shot transfer misses the attribute and is not fluent. Fine-tuned balances objectives better.}
\label{tab:ed_model_gen}
\end{table}

\subsection{Transfer experiments}

We compare three different approaches to perform attribute control anchored in this new dataset.

\paragraph{Training from scratch}
The \textsc{EmpatheticDialogues} dataset has only 25k situation descriptions, and is therefore too small to allow for successful training of the transfer architecture from scratch. To show this, we perform training exactly as in the previous section, but using only data from the 25k situation descriptions. Results in Table~\ref{tab:ed_auto_metrics} show that the system learns adequate attribute control, but ignores the source sequence.

\begin{table}
    \centering
    \begin{small}
    \begin{tabular}{p{2.5 cm}rrrr}
\toprule
&\multicolumn{2}{c}{Classification}&&\\
\cmidrule{2-3}
 & Target & Source & self-BLEU & PPL \\
 \midrule
Identity & 1.4 & 56.5 & 100.0 & 96.6 \\
Target attr. sample & 77.8 & 0.7 & 0.0 & 94.8 \\
\midrule
Scratch & 29.1 & 2.6 & 0.7 & 35.8 \\
Zero-shot & 3.6 & 30.2 & 62.0 & 135.6 \\
Fine-tuned & 33.7 & 12.4 & 33.9 & 79.2 \\
\bottomrule    
\end{tabular}
    \end{small}
    \caption{Automated metrics for transfer to attributes from the \textsc{EmpatheticDialogues} dataset.
    Metrics and baselines (top two rows) are the same as in Table~\ref{tab:none_held_out_auto_metrics}.
    Scratch:  the style transfer architecture is trained from scratch, using only the 25k situations from the \textsc{EmpatheticDialogues} dataset. The architecture learns to transfer to reasonable accuracy, but the self-BLEU scores are near zero, showing that the source content is nearly ignored. 
    Zero-shot: the transfer architecture is pre-trained to
    transfer sentiments on millions of examples from the third-party Reddit dump, and a linear mapping from the new target attributes to that embedding space is trained in a supervised way. No fine-tuning of the transfer architecture is conducted. Metrics show failure to control the target attribute or change the source sequence much, simply degrading the source sequence.
    Fine-tuned: the transfer architecture is pre-trained on the third-party Reddit dump, then fine-tuned on the \textsc{EmpatheticDialogues} situations. This achieves a much better balance between attribute control and self-BLEU. Example generations are shown in Table~\ref{tab:ed_model_gen} and Table~\ref{tab:ed_generations}. 
    }
    \label{tab:ed_auto_metrics}
\end{table}

\paragraph{Zero-shot transfer}
The ``zero-shot" approach to task transfer here requires mapping the new attribute space to the old, so as to specify the new desired targets in the embedding space understood by the model. To see if this can work without any fine-tuning, we train a logistic regression layer from the previous Reddit sentiment embedding space to the new attribute space, and use the learned attribute embeddings to specify the new target attributes. Attribute control is performed in the same way as before using a style transfer architecture trained on 20 sentiment labels (so as to allow comparing to transfer to a held-out sentiment label from the same data), but the attribute targets, the source sequences and the label classifiers are all from the \textsc{EmpatheticDialogues} dataset. This approach performs very poorly, as shown in Table~\ref{tab:ed_auto_metrics}. This is not surprising, given that the low-dimensional embedding space for the original sentiment labels is trained to represent sentiment information from conversational posts that are quite removed from the task of inferring the sentiment felt in a situation description, and may simply have lost too much information to
adequately infer the sentiment in this new context. In fact, the accuracy of the logistic regression classifier used to map the new sentiment labels to the old space is below 18\% (on the test set), compared to over 50\% achieved by a bottleneck BERT-based classifier trained on that data in raw text form.

\paragraph{Fine-tuning} Starting from the same pre-trained architecture as in the zero-shot baseline, we fine-tune the architecture on the situation descriptions from \textsc{EmpatheticDialogues}. This gives a chance for the model to adapt to the language and different framing and attribute space. Results in Table~\ref{tab:ed_auto_metrics} show that the fine-tuning reaches 
 reasonable transfer performance.
Example generations are shown in Table~\ref{tab:ed_generations}.

\begin{table}
\begin{small}
\begin{tabular}{lp{5.7cm}}
\toprule
\textbf{anxious}& \textbf{Waiting for my results} \\
anticipating & Waiting for the results to come out. \\
caring & Waiting for my grandmother. \\
joyful & Waiting for my paycheck at the end \\
prepared & Waiting for my exams \\
\midrule
\textbf{grateful}& \textbf{My grandfather invited me over and made us an awesome dinner today.} \\
hopeful & My grandfather promised to buy me a car as soon as he went on vacation. \\
jealous & My grandfather bought a car and I was pretty envious of him. \\
sad & My grandfather passed away and it was a shock. \\
\midrule
\textbf{prepared} & \textbf{I'm going overseas and i'm super ready} \\
afraid & I'm going to the doctor on Monday. I hope he does well \\
anticipating & I'm going to eat with some friends tonight. I can't wait to eat at the university. \\
confident & I'm going to get a new car this year. I just know it \\
content & I'm going overseas and i'm ready to go start my new job. \\
excited & I'm going camping next weekend. I am so stoked! \\
hopeful & I'm going to be able to get my degree next week. \\
jealous & I'm going hiking with another person who is in a relationship. \\
joyful & I'm going overseas and i'm super excited. \\
\bottomrule
\end{tabular}
\end{small}
\caption{Example generations when transferring situation descriptions from the test set of the \textsc{EmpatheticDialogues} dataset
with various source sentiment labels, to other \textsc{EmpatheticDialogues} sentiment labels. Generations are produced by the fine-tuned model in Table~\ref{tab:ed_auto_metrics}.}
\label{tab:ed_generations}
\end{table}

\section{Discussion and Conclusion}

This work has shown that taking advantage of consistent embedding spaces obtained through a separate task (in this case, supervised classification) makes it possible to achieve reasonable success with zero-shot transfer to classes that were not seen during training or even, with some fine-tuning, transfer to an altogether different attribute space.

When viewed as a method to generate controlled variations of an input text, this style transfer approach paves the way for promising data augmentation methods where an existing set of retrieval utterances could be augmented to fit specific target styles. Given that retrieval models are still performing better than generative models in conversational systems (e.g., see \citet{rashkin2019empathy}), this would allow combining the flexibility of enhanced fine-grained control with the power of retrieval models, while still escaping flaws of generative models such as blandness and repetition, similar to the retrieve-and-refine approach \citep{weston2018retrieve}.

Another promising potential use of this style transfer architecture is through the indirect, implicit definition of a style through examples: instead of requiring a label, which could lead to quantization noise when the desired attribute is not an exact match to a pre-defined attribute value, the target attribute representation can be directly inferred from an example text input that conveys the desired style. This would allow mirroring of the style of a text without labeling it, or conversely complementing it by looking at a maximally distant embedding. Our approach would also lend itself well to using un-labelled styles extracted in an unsupervised way, as long as they can be represented in a consistent embedding space.

\noindent
\bibliography{styleTransfer}
\bibliographystyle{acl_natbib}

\end{document}